\documentclass[letterpaper, 10 pt, conference]{ieeeconf}  

\IEEEoverridecommandlockouts                              

\usepackage{cite}
\usepackage{xcolor}
\usepackage{graphicx}
\usepackage{tabularx}
\usepackage{multirow}
\usepackage{booktabs}
\usepackage{makecell}
\usepackage{amsmath}
\usepackage{threeparttable}

\title{\LARGE \bf
A Simulated Experiment to Explore Robotic Dialogue Strategies for People with Dementia
}

\author{Fengpei Yuan$^{1}$, Amir Sadovnik$^{2}$, Ran Zhang$^{3}$, Devin Casenhiser$^{4}$, Eun Jin Paek$^{4}$,\\ Si On Yoon$^{5}$ and Xiaopeng Zhao$^{1}$
\thanks{*This work was not supported by any organization}
\thanks{$^{1}$The authors are with the Department of Mechanical, Aerospace and Biomedical Engineering, University of Tennessee, Knoxville, TN, USA
        {\tt\small fyuan6@vols.utk.edu, xzhao9@utk.edu}}%
\thanks{$^{2}$Amir Sadovnik is with the Department of Electrical Engineering and Computer Science, University of Tennessee,
        Knoxville, TN, USA
        {\tt\small asadovnik@utk.edu}}%
\thanks{$^{3}$Ran Zhang is with the Department of Electrical and Computer Engineering, Miami University,
        Oxford, OH, USA
        {\tt\small zhangr43@miamioh.edu}}
\thanks{$^{4}$The authors are with the Department of Audiology and Speech Pathology, University of Tennessee Health Science Center,
        Knoxville, TN, USA
        {\tt\small dcasenhi@uthsc.edu, epaek@uthsc.edu}}
\thanks{$^{5}$Si On Yoon is with the Department of Communication Sciences and Disorders, University of Iowa, 
        Iowa City, IA, USA
        {\tt\small sion-yoon@uiowa.edu}}
}

\begin{document}

\maketitle
\thispagestyle{empty}
\pagestyle{empty}

\begin{abstract}

People with Alzheimer's disease and related dementias (ADRD) often show the problem of repetitive questioning, which brings a great burden on persons with ADRD (PwDs) and their caregivers. Conversational robots hold promise of coping with this problem and hence alleviating the burdens on caregivers. In this paper, we proposed a partially observable markov decision process (POMDP) model for the PwD-robot interaction in the context of repetitive questioning, and used Q-learning to learn an adaptive conversation strategy (i.e., rate of follow-up question and difficulty of follow-up question) towards PwDs with different cognitive capabilities and different engagement levels. The results indicated that Q-learning was helpful for action selection for the robot. This may be a useful step towards the application of conversational social robots to cope with repetitive questioning in PwDs.

\end{abstract}

\section{INTRODUCTION}

According to the World Alzheimer Report 2018 \cite{Patterson2018world}, there were 50 million people living with dementia in 2018 around the world, with one new case of dementia every 3 seconds. Alzheimer's disease is the most common form of dementia, contributing to $60-70\%$ of cases \cite{WHO2021dementia}. Due to memory impairment, persons with ADRD (PwDs) often show behavior of repetitive questioning, which can be very frustrating, tedious and exhausting to their caregivers \cite{reeve2017exploration,hamdy2018repetitive}. Worse yet, caregivers often do not receive sufficient training on how to communicate with people with ADRD and do not have the time to learn the best communication methods to create a strong relationship \cite{kuwamura2016can,pou2020conversational}. Especially in senior homes, many PwDs do not receive meaningful daily interactions.
Social robots, an emerging common assistive technology to support ADRD care \cite{kruse2020evaluating}, could play a role of conversational companion and have conversations with PwDs whenever required (e.g., when a PwD shows repetitive questioning), with the advantages of high repeatability and no complaints and no fatigue \cite{taheri2015clinical}. Meaningful daily communication with a robot, as indicated in previous studies, may help reduce symptoms of ADRD and improve quality of life and independence of people with ADRD \cite{kuwamura2016can,pou2020conversational,tsiakas2016adaptive,yamazaki2019conversational,shibata2012therapeutic}.
Wang \cite{wang2017robots} found that having a robot to talk to can decrease the feelings of loneliness, and a participant even noted that ``it was nice to have someone to talk to, and the home did not feel so empty''.

On the other hand, adaptive robot interactions are necessary to provide a comfortable and effective interactions with target users \cite{cruz2018towards}, i.e., PwDs in our study.
An adaptive dialogue system would facilitate meaningful, effective communication and a more trusting relationship between the PwD and robot \cite{pou2020conversational,tsiakas2016adaptive}.
For example, Rudzicz et al. confirms that the entire communication system will be more effective if the person and mental state is taken into account \cite{rudzicz2015speech}. The adaptive (or autonomous) behaviors in robots have been investigated using the technique of Wizard of Oz (WoZ), where the robot is usually controlled by a human operator \cite{cruz2017semi,cruz2018strategies}. However, WoZ has been demonstrated to not be a sustainable technique in long term \cite{esteban2018proceedings}. WoZ may be sufficient for narrow task domains and very specific user interactions, but it is limited in terms of flexibility and adaptivity to different individual users \cite{hemminahaus2017towards}.
Particularly regarding the population of people with ADRD, each individual may have different personality, preference and cognitive abilities \cite{kobayashi2019effects}, and show time-varying behaviors, emotions (e.g., behavioral and psychological symptoms of dementia, BPSD) \cite{baharudin2019associations}, and personality \cite{islam2019personality} in both short and long term, as well as time-decreasing cognitive capabilities. 

The limitations of WoZ can be compensated by using adaptive algorithms such as reinforcement learning (RL), which enables robots to learn from the interaction with the environment (e.g., users) and makes it possible to adapt and optimize robotic policies to different individual users. It has been applied in some studies for dialogue management.
For example, Cuay{\'a}huitl used deep reinforcement learning to perform action selection from raw text for the context of restaurant \cite{cuayahuitl2017simpleds}. The state space was defined as word-based features, and the action space included $35$ dialogue actions in response to users' intentions.
In another study \cite{magyar2019autonomous} of using RL to learn a conversation strategy for autonomous robotic dialogue system for people with dementia, the authors designed state space as the robot’s internal motivation (closely associated with user's motivation) and previously selected action. Their action space was represented by three types of robot's action: short response (simple agreement/encouragement), long response (question) and topic change. The results showed the robot was capable of maintaining conversations with seniors for at least 20 mins.

Regarding the specific context of repetitive questioning in PwDs, the conversational robot would be expected to not only answer those repetitive questions for PwDs, but also further communicate with PwDs to distract their attention from the repetitive behaviors and stimulate their cognitive activities. Asking appropriate questions proactively by the robot can be a good approach to start conversations with PwDs. However, the question difficulty level (e.g., closed-ended vs open-ended questions) must be taken into careful consideration during communication with PwDs \cite{Alzheimer2021communication}. Inspired by previous relevant researches \cite{csikszentmihalyi2000beyond,tapus2009use}, asking questions with optimal difficulty level with respect to PwD's cognitive capability may help engage PwD and keep them interested in interacting with the robot, and maximally stimulate their cognitive activities.

Our long-term goal is to build a cost-effective, adaptive conversational robot, as a complement of caregivers, to cope with the problem of repetitive questioning in PwDs by answering their questioning and asking follow-up questions to distract their attention from repetitive behaviors. In this paper, we proposed a general RL framework, partially observable Markov Decision Process (POMDP) model, to firstly explore an adaptive conversation strategy in terms of the frequency of asking a follow-up question and the question difficulty level for a robot to communicate with different PwDs under the context of repetitive questioning. The RL agent adapted its policy merely based on relevance of user's response, which is easy to obtain during human-robot interaction. The learning performances of the RL model were evaluated and discussed.


\section{METHOD}

\begin{figure}[h]
    \centering
    \includegraphics[width=5.5cm]{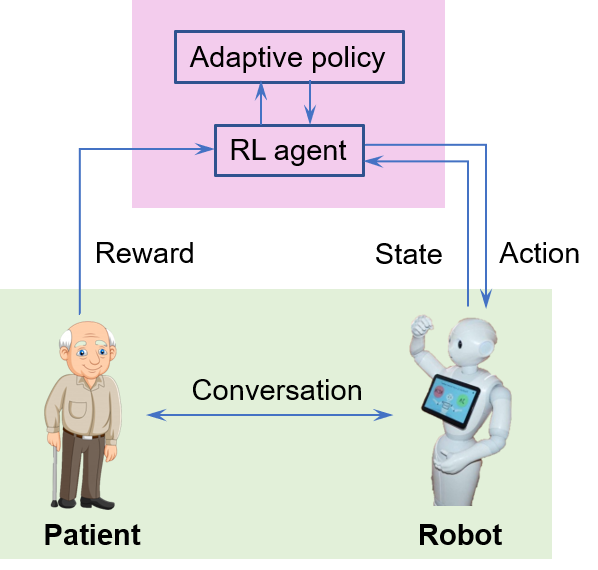}
    \caption{The framework of PwD-robot dialogue}
    \label{Fig_RLFramework}
\end{figure}

The technique of reinforcement learning is used to learn from the PwD-robot conversation (Fig.~\ref{Fig_RLFramework}) and investigate the optimal policy for a robot. As a supplement of human caregiver towards PwD's repetitive questioning, the robot should be able to always answer the questions asked by the PwD, and propose some follow-up questions to distract PwD from the repetitive behaviors and also to stimulate their cognitive activities. However, if a follow-up question is too challenging, difficult or complicated to the PwD, the PwD may not respond to it at all. Therefore, the robot should also be able to identify if the follow-up question is too difficult for the PwD, and adapt the difficulty level to users with different cognitive capabilities. Therefore, the optimal policy maps the PwD-robot interaction/dialogue to the robot's follow-up rate and difficulty level of follow-up question, so that there will be frequent conversation between the PwD and the robot, together with the brain activities in people with ADRD being maximally stimulated.

\subsection{Definition of Reinforcement Learning}

According to the aforementioned goals for PwD-robot dialogue, we define the key elements of POMDP model as follows\cite{sutton2018reinforcement}:
\begin{itemize}
    \item State space: A state is defined according to a user's situation during a conversational context. There are five potential states: the user asking a question, denoted by $Q$ or $Question$, the user's question being simply answered by robot without following-up, denoted by $NF$, the user providing relevant response to the robot's follow-up question, denoted by $RR$, the user providing irrelevant response to the robot's follow-up question, denoted by $IR$, and the user providing no response to robot's follow-up question, denoted by $NR$. Noticeably, underlying the states $RR$, $IR$, and $NR$, there are always processes of the user's question being answered by the robot and further the robot asking a follow-up question. 
    \item Action: An action includes two elements: the follow-up rate (i.e., the probability of the robot asking a question after answering PwD's question) for the robot and the difficulty level of a follow-up question. For simplification, currently the optional follow-up rate is $0.1, 0.4, 0.7$ or $1.0$. And there are three question difficulty levels: easy, moderately difficult, and difficult. Table \ref{Table_Question_Difficulty} shows examples of follow-up questions with different difficulty levels.
    \begin{table}[h]
    \centering
    \caption{Examples of follow-up questions by a robot}
    \label{Table_Question_Difficulty}
    \begin{tabular}{c|c}
       \toprule
       Question Difficulty  & Example \\
       \midrule
        Easy & ``Would you like some tea?''\\
        Moderate & ``What would you like to drink?''\\
        Difficult & ``What do you think about this tea?''\\
        \bottomrule
    \end{tabular}
\end{table}
    \item Reward: According to our goals for a social robot in the situation of repetitive questioning, i.e., answering PwDs' repetitive questions, asking follow-up questions to distract PwDs from repetitive questioning and to promote daily conversation, and avoiding too difficult follow-up questions towards PwDs, we built our immediate reward function as following,
    \begin{equation*}
        Reward =
  \begin{cases}
    0, & \text{if } $Q$ \rightarrow $NF$\\
    +1\times$QuestionDifficulty$, & \text{if } $Q$ \rightarrow $RR$\\
    +0.5, & \text{if } $Q$ \rightarrow $IR$\\
    -0.2\times$QuestionDifficulty$, & \text{if } $Q$ \rightarrow $NR$
  \end{cases}
    \end{equation*}
    where the variable $QuestionDifficulty$ can be $1$, $2$, or $3$, separately corresponding to easy, moderately difficult, and difficult follow-up questions. The expression $Q$, $NF$, $RR$, $IR$, and $NR$ represent the possible states in the MDP model. The symbol $\rightarrow$ indicates the state transition. Although meaning conversation (i.e., relevant responses from PwD to robot's questions) is the most suggested, irrelevant responses are still meaningful to PwDs. Thus, a slightly positive reward (i.e., $+0.5$) was assigned to $Q\rightarrow IR$.
\end{itemize}

\begin{figure}[h]
    \centering
    \includegraphics[width=8cm]{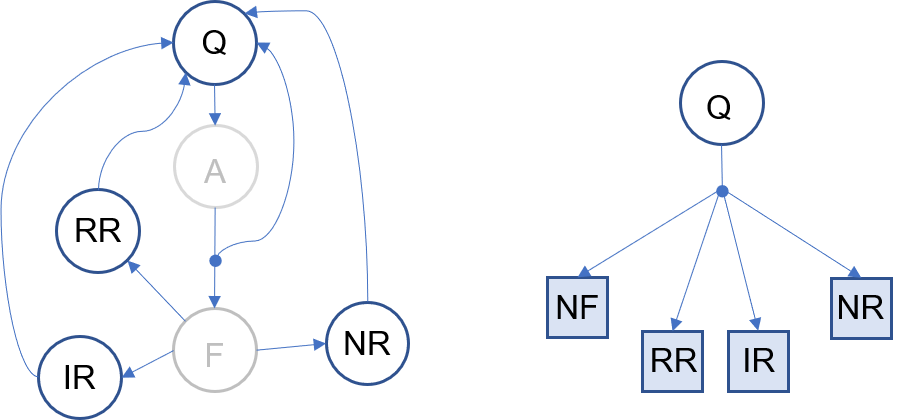}
    \caption{The MDP diagram in one episode of PwD-robot dialogue interaction}
    \label{fig_MDPDiagram}
\end{figure}

The PwD-robot dialogue is a complicated interactive task, which can be continuing (i.e., the left diagram in Fig. \ref{fig_MDPDiagram}) or episodic tasks \cite{sutton2018reinforcement}. In this paper, we started our simulated exploration study with the most simple modelling of situation, one-step episodic task, as illustrated in the right diagram of Fig. \ref{fig_MDPDiagram}. Each episodic task starts by a PwD asking a repetitive question, that is, the state $s=Q$, and ends with a terminal state, i.e., $s=NF$, $RR$, $IR$, or $NR$.  

\subsection{Simulated Person with ADRD}

Because the real-world samples of PwD-robot conversational interaction is costly in terms of time and labor, we start from simulated PwD-robot dialogue using RL modelling, which hopefully provides us insight into real-world PwD-robot dialogue in next step. A simulated individual with ADRD during the PwD-robot conversation is characterized by their response to the robot's follow-up question. The response can be categorized into three types of response: relevant, irrelevant, and no response to the follow-up question. Therefore, an individual is characterized by the relevant response rate $P_{Rresp}$ and irrelevant response rate $P_{IRresp}$. Notice here the rate of no response $P_{Nresp}$ is dependent on $P_{Rresp}$ and $P_{IRresp}$. The sum of these three variables is always $1$.

We assume that an individual's relevant and irrelevant response rates are influenced by the individual cognitive capability and their engagement, as well as the question difficulty level. Engagement of a person with ADRD is defined as the act of being occupied or involved with an external stimulus \cite{cohen2009engagement}, i.e., the conversational robot in our case. The engagement of an individual with ADRD can be influenced by robot attributes and the individual's attributes (e.g., personality and preference) \cite{cruz2020social}. There are three basic rules to create models of persons with ADRD. Firstly, with the same engagement level, a person with a lower cognitive capability will show a lower relevant response probability. Also, we expect a person with lower cognitive capability will have a higher irrelevant response probability except the most severe PwD who will have extremely low response (both relevant and irrelevant) to all questions. Secondly, we assume that an individual with a lower engagement level will have a lower response rate (i.e., the sum of $P_{Rresp}$ and $P_{IRresp}$) and thus a greater no response rate, $P_{Nresp}$. The specific effect on $P_{Rresp}$ and $P_{IRresp}$ is dependent on individuals. Therefore, the effects on individual response rate $P_{Rresp}$ and $P_{IRresp}$ are random in our simulation. 

Thirdly, given a more difficult follow-up question, the same user with the same engagement is expected to show a lower relevant response rate $P_{Rresp}$ and a higher no response rate $P_{Nresp}$. Naturally, it takes people with ADRD more cognitive workload to answer a more difficult follow-up question. For example, PwDs only need to answer "yes" or "no" to the easy question in Table \ref{Table_Question_Difficulty}. Comparatively, to answer the difficult question in this table, PwDs need to think more to understand the question and express their opinions.

During our simulation, there are four basic simulated users, each with three different levels of engagement, i.e., high, medium, and low. The parameters of the simulated User $1-4$ with different engagement levels are listed in Table \ref{Table_Simulated_User1}-\ref{Table_Simulated_User4}. User $1$, $2$, $3$ and $4$, correspond to older adults without cognitive impairment, with mild cognitive impairment, moderate dementia, and severe dementia, respectively.
\begin{table}[h]
    \centering
    \caption{Parameters of simulated user 1 without cognitive impairment}
    \label{Table_Simulated_User1}
    \begin{tabular}{l|l|l|l|l}
  \toprule
  \makecell[l]{Engage-\\ment} & \makecell[l]{Question\\difficulty} & $P_{Rresp}$ & $P_{IRresp}$ & $P_{Nresp}$ \\
  \midrule
  \multirow{3}{*}{High} & Easy & $1$ & $0$ & $0$ \\
  {} & Moderate & $1$ & $0$ & $0$ \\
  {} & Difficult & $1$ & $0$ & $0$ \\
  \hline
  \multirow{3}{*}{Medium} & Easy & $0.95$ & $0$ & $0.05$ \\
  {} & Moderate & $0.92$ & $0$ & $0.08$ \\
  {} & Difficult & $0.90$ & $0$ & $0.10$ \\
  \hline
  \multirow{3}{*}{Low} & Easy & $0.90$ & $0$ & $0.10$ \\
  {} & Moderate & $0.88$ & $0$ & $0.12$ \\
  {} & Difficult & $0.85$ & $0$ & $0.15$ \\
  \bottomrule
\end{tabular}
\end{table}
\begin{table}[h]
    \centering
    \caption{Parameters of simulated user 2 with mild cognitive impairment}
    \label{Table_Simulated_User2}
    \begin{tabular}{l|l|l|l|l}
  \toprule
  \makecell[l]{Engage-\\ment} & \makecell[l]{Question\\difficulty} & $P_{Rresp}$ & $P_{IRresp}$ & $P_{Nresp}$ \\
  \midrule
  \multirow{3}{*}{High} & Easy & $0.9$ & $0.1$ & $0$ \\
  {} & Moderate & $0.86$ & $0.14$ & $0$ \\
  {} & Difficult & $0.82$ & $0.18$ & $0$ \\
  \hline
  \multirow{3}{*}{Medium} & Easy & $0.83$ & $0.11$ & $0.06$ \\
  {} & Moderate & $0.75$ & $0.15$ & $0.10$ \\
  {} & Difficult & $0.68$ & $0.20$ & $0.12$ \\
  \hline
  \multirow{3}{*}{Low} & Easy & $0.75$ & $0.14$ & $0.11$ \\
  {} & Moderate & $0.65$ & $0.16$ & $0.19$ \\
  {} & Difficult & $0.50$ & $0.18$ & $0.32$ \\
  \bottomrule
\end{tabular}
\end{table}
\begin{table}[h!]
    \centering
    \caption{Parameters of simulated user 3 with moderate dementia}
    \label{Table_Simulated_User3}
    \begin{tabular}{l|l|l|l|l}
  \toprule
  \makecell[l]{Engage-\\ment} & \makecell[l]{Question\\difficulty} & $P_{Rresp}$ & $P_{IRresp}$ & $P_{Nresp}$ \\
  \midrule
  \multirow{3}{*}{High} & Easy & $0.70$ & $0.20$ & $0.10$ \\
  {} & Moderate & $0.63$ & $0.22$ & $0.15$ \\
  {} & Difficult & $0.50$ & $0.23$ & $0.27$ \\
  \hline
  \multirow{3}{*}{Medium} & Easy & $0.60$ & $0.21$ & $0.19$ \\
  {} & Moderate & $0.50$ & $0.25$ & $0.25$ \\
  {} & Difficult & $0.30$ & $0.20$ & $0.50$ \\
  \hline
  \multirow{3}{*}{Low} & Easy & $0.35$ & $0.15$ & $0.50$ \\
  {} & Moderate & $0.20$ & $0.13$ & $0.67$ \\
  {} & Difficult & $0.08$ & $0.10$ & $0.82$ \\
  \bottomrule
\end{tabular}
\end{table}
\begin{table}[h!]
    \centering
    \caption{Parameters of simulated user 4 with severe dementia}
    \label{Table_Simulated_User4}
    \begin{tabular}{l|l|l|l|l}
  \toprule
  \makecell[l]{Engage-\\ment} & \makecell[l]{Question\\difficulty} & $P_{Rresp}$ & $P_{IRresp}$ & $P_{Nresp}$ \\
  \midrule
  \multirow{3}{*}{High} & Easy & $0.04$ & $0.08$ & $0.88$ \\
  {} & Moderate & $0.02$ & $0.08$ & $0.90$ \\
  {} & Difficult & $0.01$ & $0.04$ & $0.95$ \\
  \hline
  \multirow{3}{*}{Medium} & Easy & $0.02$ & $0.05$ & $0.93$ \\
  {} & Moderate & $0.01$ & $0.04$ & $0.95$ \\
  {} & Difficult & $0.005$ & $0.02$ & $0.975$ \\
  \hline
  \multirow{3}{*}{Low} & Easy & $0.01$ & $0.04$ & $0.95$ \\
  {} & Moderate & $0$ & $0.02$ & $0.98$ \\
  {} & Difficult & $0$ & $0.01$ & $0.99$ \\
  \bottomrule
\end{tabular}
\end{table}

\subsection{Experiments}

We trained our reinforcement learning model for our four simulated users with three different engagement levels. We applied the technique of Off-policy Q-learning to solve the MDP problem, with the $\epsilon$-greedy policy ($\epsilon=0.1$) and a learning rate of $0.03$. We evaluated the performance of our model with the metrics proposed by \cite{tsiakas2016adaptive}, including the average return per epoch, the starting state value $V(Question)$ in each epoch, and the sum of Q-value updates during each epoch. Here, due to our definition of one-step episodic task, we only need to consider the starting state value in the starting state, $Question$ (i.e., the PwD asking repetitive question). The start state value $V(Question)$ indicates the expected return. Each epoch is composed of $150$ episodes. Furthermore, we analyzed the performance of Q-learning by comparing the learning results using randomized action selection.

\begin{figure*}[h]
    \centering
    \includegraphics[width=14cm]{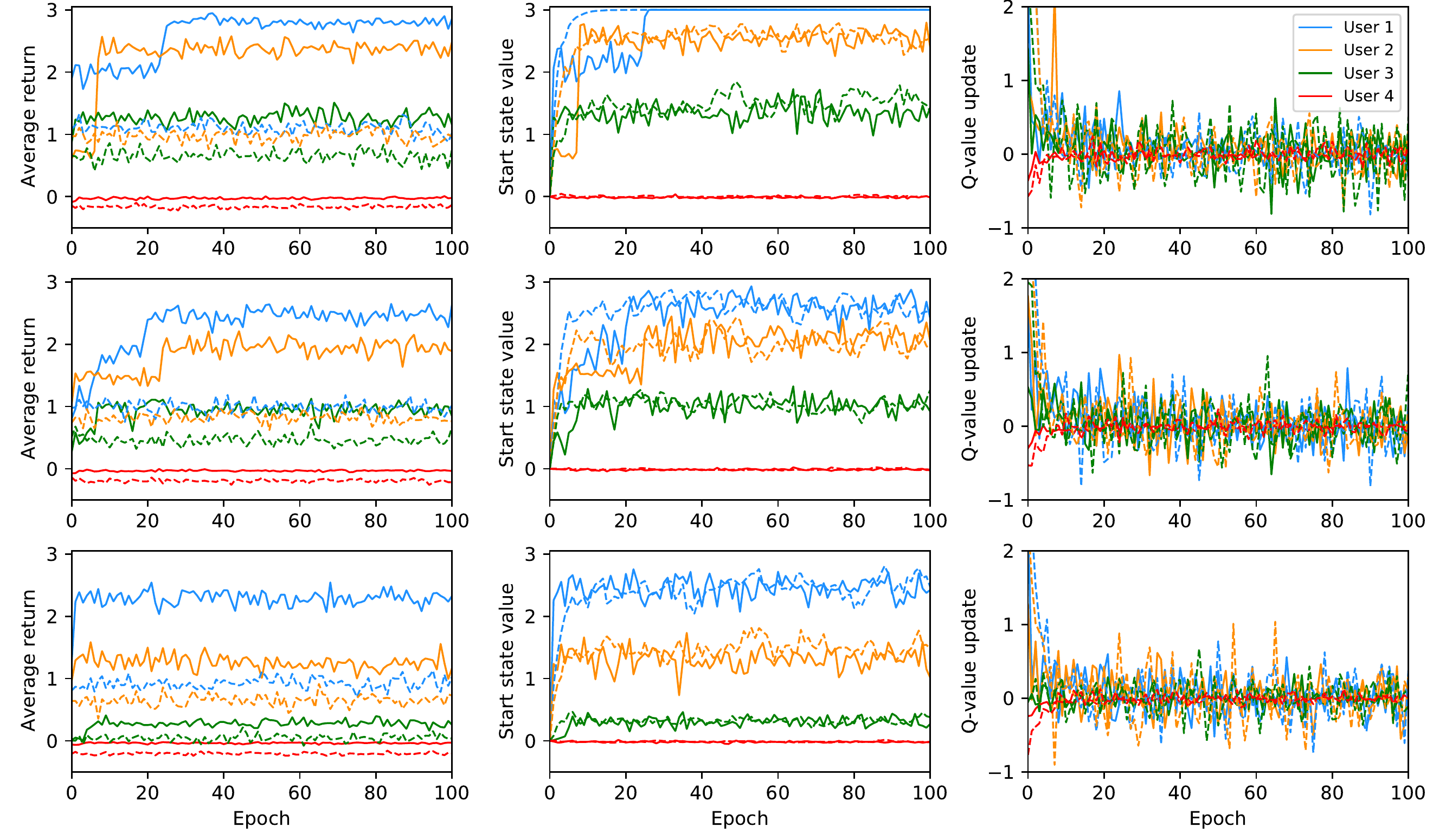}
    \caption{The learning results for four users with high (first panel), medium (second panel), and low (third panel) engagement. In each sub-figure, the solid and dashed curves represent the learning results by Q-learning and random action selection model, respectively.}
    \label{fig_InitialLearningResults_G_V_Q}
\end{figure*}




\section{Results}

Based on the evaluation metrics, we observed that in all cases (i.e., User $1-4$ with different engagement level) our RL model converged within $30$ epochs. Therefore, we only showed the learning processes of the model during epoch $1-100$ in Fig. \ref{fig_InitialLearningResults_G_V_Q}. In this figure, the first, second, and third column represent the average return obtained per epoch, the state value $V(Question)$ at the beginning of each epoch, and the sum of Q-value updates over all $150$ episodes in each epoch, separately. And the upper, middle, and lower row correspond to the learning process of all users with high, medium, and low engagement, respectively. The solid and dashed curves with the same color in each sub-figure represent the learning results for the same user using Q-learning and random action selection, respectively. A user with a higher engagement level is associated with greater average return, compared to the same user with lower engagement level. With the same engagement level, the converged average return in User $1$ and $2$ was obviously greater than that in User $3$ and $4$. The optimal policies learned by the RL agent for different users with different engagement levels were listed in Table \ref{Table_OptimalPolicy}.

\begin{table}[h]
    \centering
    \caption{Optimal policy suggested by the Q-learning}
    \label{Table_OptimalPolicy}
    \renewcommand{\arraystretch}{1.2}
    \begin{tabular}{l|l|l|l|l}
        \toprule
        Engagement & User $1$ & User $2$ & User $3$ & User $4$ \\
        \midrule
        High & $[1.0,D]$ & $[1.0,D]$ & \makecell[l]{$[1.0,D]$ or\\$[1.0,M]$} & $[0.1,E]$\\
        Medium & $[1.0,D]$ & $[1.0,D]$ & $[1.0,M]$ & $[0.1,E]$\\
        Low & $[1.0,D]$ & \makecell[l]{$[1.0,D]$ or\\$[1.0,M]$} & $[1.0,E]$ & $[0.1,E]$\\
        \bottomrule
    \end{tabular}
    \begin{tablenotes}
    \item \textit{Note} User $1$, $2$, $3$ and $4$ represent older adults without cognitive impairment, with mild cognitive impairment, moderate dementia and severe dementia. $D=$ difficult follow-up questions; $M=$ moderately difficult follow-up questions; $E=$ easy follow-up questions.
    \end{tablenotes}
\end{table}

\section{DISCUSSION and FUTURE WORK}

In this paper, we applied the reinforcement learning method to explore the high-level strategy for the human-robot dialogue system in the situation of repetitive questioning by persons with ADRD. The strategy may allow a social robot to have an adaptive conversation with PwDs to their cognitive capability and engagement level.
From Fig. \ref{fig_InitialLearningResults_G_V_Q}, we can see that the implemented RL agent is able to learn the best policy within $30$ epochs. 
As shown in the first column of Fig. \ref{fig_InitialLearningResults_G_V_Q}, the average return curves for all users learned by Q-learning are greater than curves learned by random action selection, which indicated that Q-learning here are helpful for action selection. 
As training evolves, the average return curve by Q-learning in the first column tend to approximate the start state value $V(Question)$ curve in the second column, which was also found in previous RL modelling study \cite{tsiakas2016adaptive}.
At the same time, the total Q-value update by Q-learning per epoch becomes $0$, as shown in the third column. Noticeably, the learning processes converge with spikes, which is due to the simulation of stochastic response rate (Table \ref{Table_Simulated_User1}-\ref{Table_Simulated_User4}).  

In each block of the first two columns in Fig. \ref{fig_InitialLearningResults_G_V_Q}, as the learning curves converge, the average return and the start state value $V(Question)$ for an individual with higher cognitive capability (e.g., User $1$ and $2$) are greater than an individual with lower cognitive capability (e.g., User $1$ and $2$). This makes sense considering that an older adult without or with mild cognitive impairment is more likely to answer a question. 
Comparing the optimal policies learned by our RL agent for four types of user with the same engagement level, for example, the row of medium engagement, the agent is able to learn the the best policy for PwDs with different cognitive capabilities. More specifically, the optimal policy (i.e, follow-up rate and question difficulty) for a user with higher cognitive capability (e.g., User $1$ and $2$) is always asking the difficult follow-up question. However, towards users with lower cognitive capability, User $3$ and $4$, the RL agent separately adapts the policy to always following up with moderate-level difficult question, and following up very rarely also with easy questions.

Comparing the learning results for the same user but with different engagement level, for example, the green curve representing User $3$ in three blocks of the first column in Fig. \ref{fig_InitialLearningResults_G_V_Q}, the average return obtained per epoch decreases as the user's engagement decreases, which makes sense because an individual with lower engagement is expected to less likely join activities and conversations. Moreover, the column of User $3$ in Table \ref{Table_OptimalPolicy} shows that the question difficulty needs to decrease as an adaptation to individual's decreasing engagement level. This indicates that, although merely observing relevance of user's response, the RL agent is able to detect the change of latent variable (i.e. user's engagement level) and adaptively adjust the question difficulty level accordingly.

There are some limitation in the study. First, a state in our POMDP only considered the relevance of user's response to a robot's question. Although the current POMDP seems to adapt the policy when the latent variable (e.g., user's engagement level) changes, the inclusion of more variables associated with users, e.g., PwD's cognitive and affective states, may facilitate better learning performance. Previous studies has showed that a PwD's engagement level can be read using sensors such as camera \cite{chen2020teaching}, heart-rate sensor \cite{cruz2018towards} and electroencephalography (EEG) \cite{khedher2019tracking}. We will take into consideration of integrating these variables into the state space in a cost-effective way.
Second, we made some simplification that an individual's cognitive capability and engagement level is consistent during the whole PwD-robot dialogue, which is usually not true in real world considering the intra-individual variability and disease progression in people with ADRD. In future, we will conduct more research on this direction, to investigate the performance of our RL model for users with time-decreasing cognitive capability and time-varying engagement.
Another limitation is that our current RL model only considers the interaction between a PwD and a robot. In future, we will incorporate the intervention from their caregivers or therapist, as suggested by \cite{tsiakas2016adaptive}. The inclusion of caregivers' intervention will not only ensure the safety of PwD-robot interaction, but also may facilitate a faster optimization of policy by RL agent.

\section{CONCLUSIONS}

In this paper, we developed a general RL framework, POMDP model, to learn the conversation strategy for a robot to cope with the problem of repetitive questioning by persons with ADRD. The model allows the robot's dialogue system to adjust an appropriate rate of asking a follow-up question and the question difficulty level.
This study may allow a conversational social robot to help caregivers with PwDs' repetitive questioning behaviors. Moreover, the design of follow-up question may also distract PwDs from repetitive behaviors, which is suggested by Alzheimer's caregiving, and stimulate PwDs' brain activities through the conversation with different difficulty level. In the future, the POMDP model will be improved leveraging the multimodal sensing. 
Also, more work is needed to train the RL model with users who has time-decreasing cognitive capabilities and time-varying engagement level. Also, future study is needed to implement this model in our physical robot and investigate the performance with PwDs in real world.

\addtolength{\textheight}{-12cm}   








\bibliographystyle{IEEEtran}

\bibliography{Ref}

\end{document}